\def\year{2022}\relax
\documentclass[letterpaper]{article} 
\usepackage{aaai22}  
\usepackage{times}  
\usepackage{helvet}  
\usepackage{courier}  
\usepackage[hyphens]{url}  
\usepackage{graphicx} 
\usepackage{natbib}  

\usepackage{caption} 
\DeclareCaptionStyle{ruled}{labelfont=normalfont,labelsep=colon,strut=off} 
\usepackage{multirow}
\usepackage{xcolor}
\usepackage{amsmath}
\usepackage{amssymb}
\usepackage{colortbl}
\definecolor{grayDark}{gray}{0.95}
\definecolor{grayLight}{gray}{0.98}

\usepackage{algorithm}
\usepackage{algorithmic}

\usepackage{newfloat}
\usepackage{listings}
\floatstyle{ruled}
\newfloat{listing}{tb}{lst}{}
\floatname{listing}{Listing}
\title{Stereo Neural Vernier Caliper}
\author{
    Shichao Li\thanks{The correspondence author is Shichao Li.}\textsuperscript{\rm 1}, 
    Zechun Liu\textsuperscript{\rm 1,2}, 
    Zhiqiang Shen\textsuperscript{\rm 2,3,1},
    Kwang-Ting Cheng\textsuperscript{\rm 1}
}

\affiliations{
    \textsuperscript{\rm 1}Hong Kong University of Science and Technology \\
    \textsuperscript{\rm 2}Carnegie Mellon University,
    \textsuperscript{\rm 3}Mohamed bin Zayed University of Artificial Intelligence \\
    slicd@cse.ust.hk,
    timcheng@ust.hk
}

\begin{document}

\maketitle

\begin{abstract}
	We propose a new \emph{object-centric} framework for learning-based stereo 3D object detection. Previous studies build \emph{scene-centric} representations that do not consider the significant variation among outdoor instances and thus lack the flexibility and functionalities that an instance-level model can offer. We build such an instance-level model by formulating and tackling a \emph{local update problem}, i.e., how to predict a refined update given an initial 3D cuboid guess. We demonstrate how solving this problem can complement \emph{scene-centric} approaches in (i) building a \emph{coarse-to-fine} multi-resolution system, (ii) performing model-agnostic object location refinement and, (iii) conducting stereo 3D tracking-by-detection. Extensive experiments demonstrate the effectiveness of our approach, which achieves state-of-the-art performance on the KITTI benchmark. Code and pre-trained models are available at \url{https://github.com/Nicholasli1995/SNVC}. 
\end{abstract}

\section{Introduction}

Accurate perception of surrounding objects' 3D attributes is indispensable for autonomous driving, robot navigation, and traffic surveillance. Active range sensors such as LiDAR measures the 3D scene geometry directly to perform precise 3D localization~\cite{lang2019pointpillars, shi2019pointrcnn}. However, LiDAR sensors incur a high cost and can be limited in perception range where distant objects are only captured with very few points. On the other hand, passive sensors like cameras are inexpensive, yet the depth information is lost during the image formation process which makes 3D scene understanding a challenging inverse problem. Estimating depth from a \emph{single} RGB image is ill-posed and leads to limited 3D object detection performance~\cite{brazil2019m3d, wang2021depth, lu2021geometry}. Stereo cameras, simulating a \emph{binocular} human vision system, are the minimum sensor configuration that can exploit multi-view geometry for more reliable depth inference. Studying stereo 3D object detection (S3DOD),  thus not only is a pursuit of the vision community that aims at visual scene understanding, but also offers practical value to complement active sensors through multi-sensor fusion~\cite{MMF}.       

\begin{figure}
	\begin{center}
		\includegraphics[width=0.7\linewidth, trim=1cm 1cm 1cm 1cm]{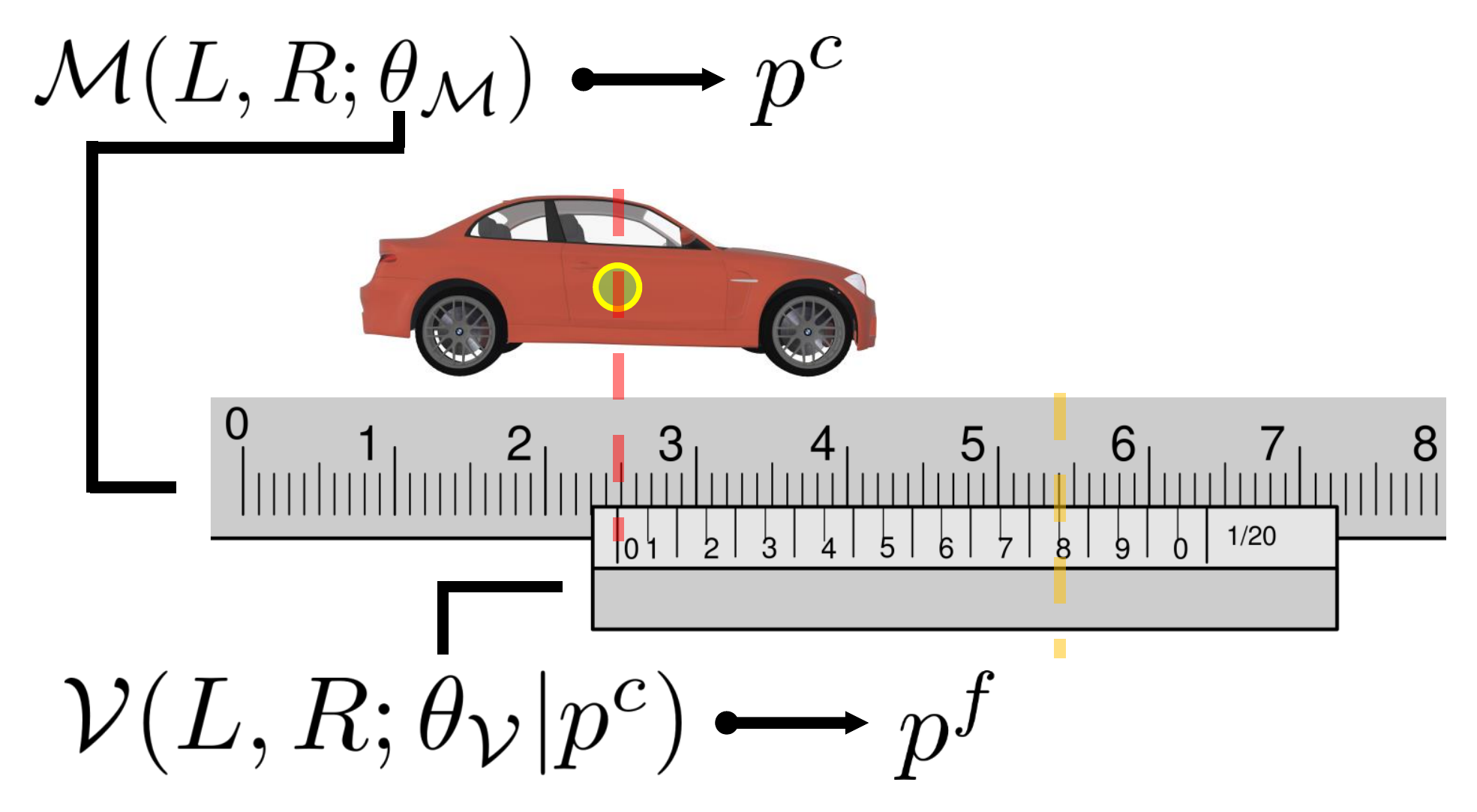}
	\end{center}
	\caption{Stereo Neural Vernier Caliper (SNVC) consists of the main scale network $\mathcal{M}$ that models a coarse global scene and the Vernier scale network $\mathcal{V}$ that models a fine local scene. $\mathcal{M}$ predicts coarse location $p^c$ while $\mathcal{V}$ \emph{takes a closer look} around $p^c$ and makes finer measurements.}
	\label{fig:teaser}
\end{figure}
Recent state-of-the-art (SOTA) S3DOD approaches take a \emph{scene-centric} view and build a data representation for the \textbf{whole} scene. We use a representative work~\cite{chen2020dsgn} as our baseline, which uses the estimated depth to build a voxel-based scene representation for object detection. In contrast to it, our study promotes an \emph{object-centric} viewpoint and explores instance-level analysis for S3DOD. The following practical considerations motivate this study, which demand the attributes of this \emph{object-centric} viewpoint that are not offered by the \emph{scene-centric} counterpart.
\begin{itemize}
	\item Cost-accuracy trade-off considering the depth variation: a distant object usually has lower-resolution (LR) feature representation compared to a nearby one, which makes it difficult to precisely recover its 3D attributes. Naively re-computing high-resolution (HR) features and using a finer voxel grid for the \textbf{whole} scene lead to intimidating computational cost and cubic memory growth. This is also unnecessary if the model performance for nearby instances already saturates. If an instance-level model is available, a multi-resolution (MR) system can be built to mitigate this problem. This system benefits from a \emph{coarse-to-fine} design which computes an LR global representation and focuses on the tiny instances with complementary HR features\footnote{Such HR features can be obtained computationally or physically with an actively zooming camera~\cite{bellotto2009distributed}.}. 
	\item Capability for efficiently handling new frames: videos are more prevalent in real-world scenarios than two static stereo images. The \emph{scene-centric} approaches need to build a scene representation for every new pair of frames. With an instance-level model, one only needs to do so for some key frames and can conduct \emph{tracking-by-detection}~\cite{andriluka2008people}, i.e., processing only a portion of new frames given region-of-interests (RoIs) implied by past detections.
	\item Flexibility in video applications: certain objects are more important in a 3D scene, e.g., a car heading towards a driver should draw more attention than a vehicle leaving the field of view. Instance-level analysis can offer the flexibility to prioritize certain regions when analyzing new frames.
\end{itemize}
To make an instance-level analysis model useful in all aforementioned scenarios, we can design it as a refinement model in a recurrent manner. Given an initial 3D bounding box guess, the model should reason about the 3D space around the cuboid guess and give an updated prediction. Note the initial guess can vary for different user scenarios, e.g., it can be a proposal from an LR global model or the prediction of the last frame. Thus our research question is \emph{how to design such an instance-level refinement model in the stereo perception setting}? We tackle this problem by designing an instance-level neural network $\mathcal{V}$ which builds a voxel-based local scene representation and scores each voxel how likely it is a potential update of an object part. $\mathcal{V}$ can be combined with a global model $\mathcal{M}$ to form an MR S3DOD framework as illustrated in Fig.~\ref{fig:teaser}. $\mathcal{M}$ performs \emph{scene-level} depth estimation and outputs coarse 3D object proposals. Conditioned on each 3D proposal, $\mathcal{V}$ further extracts HR features and refines its 3D attributes. We name this framework Stereo Neural Vernier Caliper (SNVC) since it resembles a Vernier caliper where $\mathcal{M}$ (the main scale) models the 3D scene with a coarse voxel grid while $\mathcal{V}$ (the Vernier scale) models an HR local scene conditioned on the initial guess. Compared to prior arts, our approach endows an S3DOD system with the advantages discussed above and can model fine-grained structures for important regions with a tractable number of voxels. Such ability leads to superior detection performance, especially for the hard instances, i.e., the tiny and occluded ones. This paper's contributions are summarized as:

\begin{itemize}
	\item We propose the first MR framework for voxel-based S3DOD. The new instance-level model $\mathcal{V}$ within it, to our best knowledge, is also the first HR voxel representation learning approach tackling the \emph{local update problem}.
	\item We study the transferability of $\mathcal{V}$ and demonstrate it can be used as a model-agnostic and plug-and-play refinement module for S3DOD that complements many existing \emph{scene-centric} approaches.
	\item SNVC out-performs previously published results on the KITTI benchmark for S3DOD at the date of submission (Sep 8th, 2021).
\end{itemize} 

\section{Related Work}
Our study is relevant to the following research directions while having distinct contributions.

\noindent \textbf{Learning-based 3D object detection} aims to learn a mapping from sensor input to 3D bounding box representations~\cite{3dop}. Depending on the sensor modality, two parallel lines of research are vision-based methods ~\cite{oftnet, ke2020gsnet, monopseudoliar, wang2021plume, reading2021categorical} and LiDAR-based approaches~\cite{MV3D, second, VoxelNet, fpointnet, li2021lidar}. Our approach lies in the former which does not require expensive range sensors. Compared with previous stereo vision-based studies~\cite{chen2020dsgn, div2020wstereo} that focus on global scene modeling, we deal with a different \emph{local update problem} and dedicate a new model for HR instance-level analysis. Our design can complement previous 3D object detectors that do not share the flexibility and high precision offered by our approach.  

\noindent \textbf{Instance-level analysis in 3D object detection} builds a feature representation for an instance proposal to estimate its high-quality 3D attributes. FQ-Net~\cite{liu2019deep} draws a projected cuboid proposal on an instance patch and regresses its 3D Intersection-over-Union (IoU) for location refinement. RAR-Net~\cite{liu2020reinforced} formulates a reinforcement learning framework for iterative instance pose refinement. 3D-RCNN~\cite{kundu20183d} uses instance shape as auxiliary supervision yet requires extra annotations which are not needed by our approach. Notably, all these methods only consider the monocular case and \emph{cannot} utilize stereo imagery. ZoomNet~\cite{xu2020zoomnet} and Disp R-CNN~\cite{sun2020disp} construct point-based representations for each instance proposal and require extra mask annotation during training. Such representations also lose the semantic features and are less robust for distant and occluded objects which have few foreground points. We instead propose to learn a voxel-based representation to encode both semantic and geometric features. 

\noindent \textbf{Voxel-based representation} is a classical and simple data structure encoding 3D features and is widely adopted in image-based rendering~\cite{seitz1999photorealistic} and multi-view reconstruction~\cite{vogiatzis2005multi}. Early studies utilize hand-crafted features and energy-based models to encode prior knowledge~\cite{snow2000exact}, while recent deep learning-based approaches~\cite{choy20163d, riegler2017octnet} directly learn such representations from data. Our approach learns a voxel-based neural representation for the unique \emph{local update problem} under the stereo perception setting.

\noindent \textbf{High-resolution neural networks} was recently proposed~\cite{sun2019deep} to model fine-grained spatial structure details to benefit tasks that involve precise localization. However, later studies only focus on monocular and 2D tasks~\cite{wang2020deep, cheng2020higherhrnet, li2020cascaded, Li_2021_CVPR}. This work instead studies HR representation learning for S3DOD and can build an unprecedented fine 3D spatial resolution of 3 centimeters in the real self-driving scenarios\label{key}.

\noindent \textbf{Multi-resolution volumetric representation} was previously studied for 3D shape representation~\cite{riegler2017octnet} and reconstruction~\cite{blaha2016large} where less important region (e.g., free space) is represented with coarse voxel grid and a finer voxel grid is used for regions close to object surface adaptively. The octree~\cite{laine2010efficient} was a popular choice to generate an MR partition of a 3D region. We argue that there is also a large variation of importance for different regions in S3DOD where regions near objects have a larger influence on the detection performance. We thus introduce the idea of varying resolution to voxel-based S3DOD for the first time with a new MR system.

\section{Cascaded 3D Object Detection}
SNVC can be formulated as a cascaded model represented as the set $\{\mathcal{M}, \mathcal{V}\}$ consisting of the main scale network $\mathcal{M}$ and the Vernier scale network $\mathcal{V}$. Given an RGB image pair $(\mathcal{L}, \mathcal{R})$ captured by calibrated stereo cameras, $\mathcal{M}$ builds a coarse representation of the \emph{global} 3D scene and predicts $q$ 3D bounding box proposals $\{ p_q^c \}_{q=1}^{N}$ as
\begin{equation}
\mathcal{M}(\mathcal{L}, \mathcal{R}; \theta_{\mathcal{M}}) = \{ p_q^c \}_{q=1.}^{N}
\end{equation}
Conditioned on each coarse proposal $p_q^c$, $\mathcal{V}$ deals with the \emph{local update problem} by constructing an HR local scene representation and infers an offset $\delta p_q$ to obtain the refined pose $p_q^f = p_q^c + \delta p_q$.    
\begin{equation}
\mathcal{V}(\mathcal{L}, \mathcal{R}; \theta_{\mathcal{V}}|p_q^c) = \delta p_{q.}
\end{equation}
This framework is general and does not enforce any assumption on the architecture design of $\mathcal{M}$. As we will show, one design of $\mathcal{V}$ can handle coarse predictions from different implementations of $\mathcal{M}$ and thus be \emph{model-agnostic}. In light of this, we only detail our architecture of $\mathcal{V}$ in the main text. $\mathcal{M}$ used in our final system is sketched in Fig.~\ref{fig:mainscale} whose architecture details are in the supplementary material (SM).

\section{High-resolution Instance-level Update}
In this study,  we propose to design $\mathcal{V}$ based on a three-step procedure:
\begin{itemize}
	\item Space partitioning: we build a dense voxel grid in a 3D RoI conditioned on the coarse 3D proposal where each voxel is a candidate for a part location update.
	\item Deep voxel coloring: we aggregate high-level features extracted by a deep neural network for each voxel.
	\item Voting: we extract spatial structural information from the colored voxel grid and score each voxel how likely it is a location update of an object part. The pose update is then estimated via a vote based on the predicted object part locations and confidences.
\end{itemize}

The first step builds a very dense local 3D scene representation aiming at high-precision 3D object localization. In step two we study two strategies that correspond to two user scenarios and lead to two architecture variants. Our strategy in step three utilizes several pre-defined cuboid parts for robust inference considering occlusion. The following sub-sections describe each step in detail.  

\begin{figure}
	\begin{center}
		\includegraphics[width=0.9\linewidth, trim=1cm 1cm 1cm 1cm]{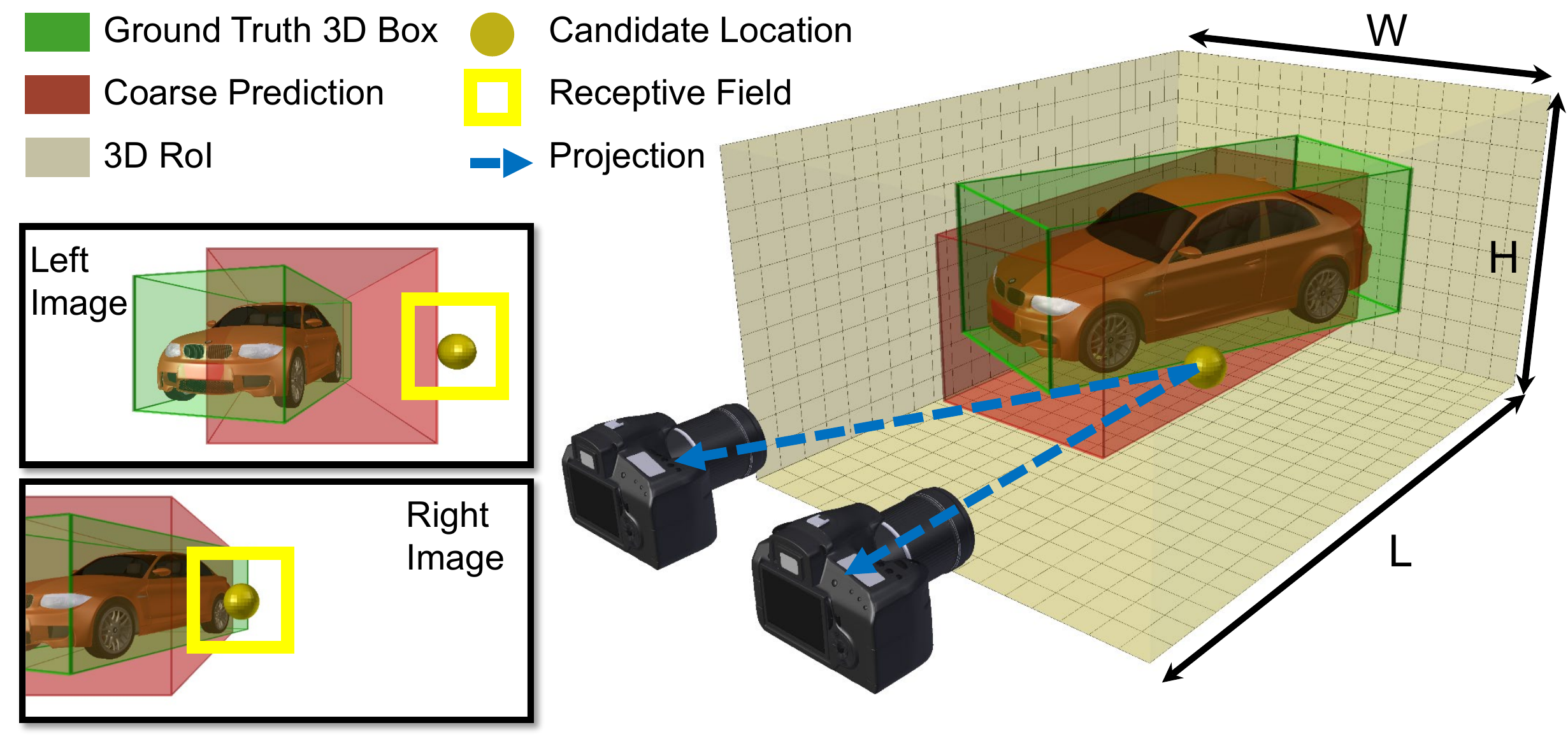}
	\end{center}
	\caption{Illustration of the \emph{local update problem}. A 3D region-of-interest (brown) is defined based on the coarse proposal (red). Each candidate (yellow ball) aggregates high-level stereo visual features. Such high-resolution volumetric features are used to infer the ground truth 3D bounding box (green). Best viewed in color. A coarse grid is drawn for visualization while the real grid is much finer.}	
	\label{candidates}
\end{figure}

\noindent \textbf{Candidate generation.}
Given a coarse 3D bounding box prediction $p_q^c$ represented as a 7-tuple $p_q^c = (x_q, y_q, z_q, h_q, w_q, l_q, \theta_q)$, where $(x_q, y_q, z_q)$, $(h_q, w_q, l_q)$ and $\theta_q$ denotes its translation, size (height, width, length) and orientation respectively. These quantities are represented in the camera coordinate system (CCS). We define a 3D RoI $r_q$ around $p_q^c$, within which $\mathcal{V}$ needs to predict the instance pose update. We use a cuboid RoI for convenience and represent $r_q$ as $(x_q, y_q, z_q, H, W, L, \theta_q)$. $r_q$ has the same 4-D pose $(x_q, y_q, z_q, \theta_q)$ as $p_q^c$ and a pre-defined range $(H, W, L)$. The \emph{Vernier scale} is a fine partition of $r_q$ represented as a 3D voxel grid $\{d_{i,j,k}\}_{i=1,j=1,k=1}^{N_H,N_W,N_L}$ with $N_H$, $N_W$, and $N_L$ voxels uniformly sampled along the height, width and length directions respectively. Each voxel $d_{i,j,k}$ is a candidate for precise location update. Fig.~\ref{candidates} illustrates a $r_q$ and its corresponding candidates. 

We represent $r_q$ with the camera coordinates of its center and the eight corners by applying a homography encoding translation and rotation as

\begin{equation}
\text{H}_{4 \times 4}
\begin{bmatrix}
\vline \\
\text{O}_{3 \times 9} \\
\vline \\
\textbf{1}
\end{bmatrix}
=
\begin{bmatrix}
cos\theta_q & 0 & sin\theta_q & x_q \\
0 & 1& 0& y_q \\
-sin\theta_q & 0& cos\theta_q& z_q \\
0 & 0& 0& 1
\end{bmatrix}
\begin{bmatrix}
\vline \\
\text{O}_{3 \times 9} \\
\vline \\
\textbf{1}
\end{bmatrix},
\end{equation}
where $\text{O}_{3 \times 9}$ represent the 9 parts in the object coordinate system as
\begin{equation}
\label{parts}
\begin{bmatrix}
0& \frac{L}{2}& \frac{L}{2}& \frac{L}{2}& \frac{L}{2}& -\frac{L}{2}& -\frac{L}{2}& -\frac{L}{2}& -\frac{L}{2}\\[1.5pt]
0& -\frac{H}{2}& \frac{H}{2}& -\frac{H}{2}& \frac{H}{2}& -\frac{H}{2}& \frac{H}{2}& -\frac{H}{2}& \frac{H}{2} \\[1.5pt]
0& \frac{W}{2}& \frac{W}{2}& -\frac{W}{2}& -\frac{W}{2}& \frac{W}{2}& \frac{W}{2}& -\frac{W}{2}& -\frac{W}{2}
\end{bmatrix}.
\end{equation}
Similarly, $d_{i,j,k}$ can be represented in the CCS as 
\begin{equation}
d_{i,j,k} = 
\text{H}_{4 \times 4}
\begin{bmatrix}
-N_L \Delta L / 2 + (k-1)\Delta L\\
-N_H \Delta H / 2 + (i-1)\Delta H \\
N_W \Delta W / 2 - (j-1)\Delta W \\
1
\end{bmatrix},
\end{equation}
where $[ \frac{-N_L \Delta L}{2},  \frac{-N_H \Delta H}{2}, \frac{N_W \Delta W}{2}]$ is the left-back-top corner of the 3D grid, and $[\Delta L, \Delta H, \Delta W] = [\frac{L}{N_L}, \frac{H}{N_H}, \frac{W}{N_W}]$ gives the grid resolution. In experiments, we specify $\Delta L, \Delta H, \Delta W$ as 3, 10, and 3 centimeters respectively and $[N_L, N_H, N_W] = [192, 32, 128]$. Such 3-cm resolution is in contrast to a 20-cm or even coarser resolution used in our $\mathcal{M}$ and previous studies~\cite{chen2020dsgn, li2020rts3d, div2020wstereo}.

\noindent \textbf{Feature aggregation.} We design two types of feature aggregation strategies depending on whether one can reuse features computed by $\mathcal{M}$. These two strategies result in two model variants ($\mathcal{V}$-A and $\mathcal{V}$-S). 

For the $\mathcal{V}$-A (\emph{model-agnostic}) models, we assume no knowledge of the architecture of $\mathcal{M}$ and do not reuse its features. This type of model is useful for dealing with proposals from different 3DOD models or handling new frames that have no features computed yet. In this case, we aggregate feature for each candidate from the left/right visual features $\mathcal{L}^F/\mathcal{R}^F$, which are obtained from a fully convolution network $\mathcal{L}^F/\mathcal{R}^F = \mathcal{N}^{2D}(\mathcal{L}/\mathcal{R};\theta_{\mathcal{N}})$. Specifically, the feature vector for $d_{i,j,k}$ is
\begin{equation}
F_{i,j,k} = \mathcal{W}(\mathcal{L}^F, K_ld_{i,j,k}) \oplus \mathcal{W}(\mathcal{R}^F, K_rd_{i,j,k}),
\end{equation}
where $K_l, K_r$ are the intrinsic parameters of the left/right cameras and $\mathcal{W}^F$ is a warping function and $\mathcal{W}(\mathcal{L}^F, K_ld_{i,j,k})$ extracts the feature at $K_ld_{i,j,k}$ from the corresponding location on the feature maps $\mathcal{L}^F$. $\mathcal{W}$ is implemented as bi-linear interpolation. 

For the $\mathcal{V}$-S (\emph{model-specific}) models, we assume that $\mathcal{M}$ also constructs a volumetric scene representation (e.g., a global cost-volume $\mathcal{C}_{g}$). This case happens when one builds a two-stage voxel-based detector where $\mathcal{V}$ has access to pre-computed features. Here we sample $F_{i,j,k}$ from such features as $F_{i,j,k} = \mathcal{W}(\mathcal{C}_{g}, d_{i,j,k})$ to save computation.  

After aggregation, the grid is colored with high-level visual features. This is in contrast to the low-level pixel intensity used in the classical voxel coloring study~\cite{seitz1999photorealistic}. In addition, all voxels are colored instead of coloring only the consistent voxels~\cite{seitz1999photorealistic} because the Lambertian scene assumption does not hold. Subsequently a network $\mathcal{N}^{3D}$ processes the aggregated volumetric representation and predicts the output detailed below.

\begin{figure*}[t]
	\begin{center}
		\includegraphics[width=1\linewidth, trim=0cm 0cm 0cm 0cm]{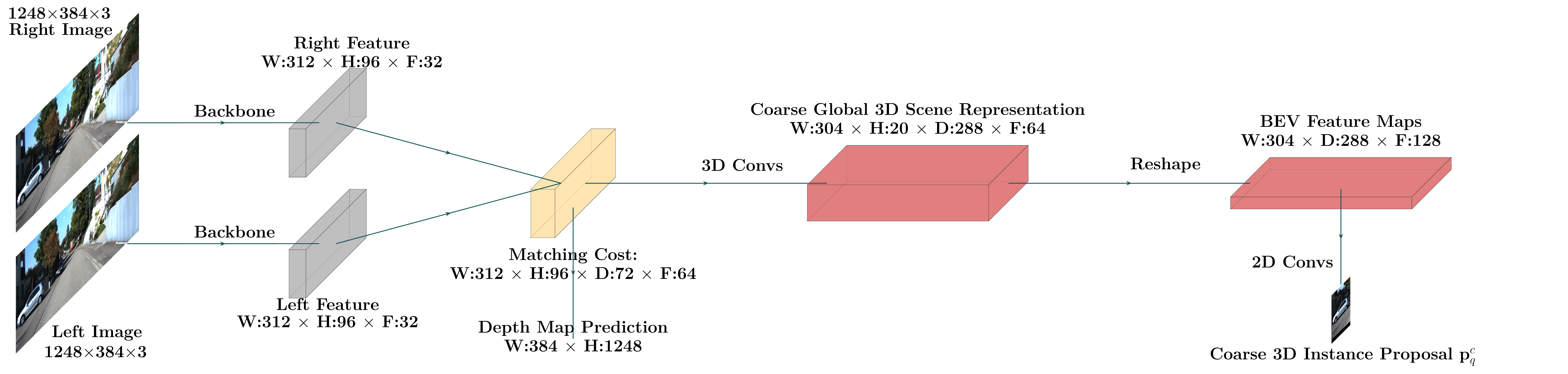}
	\end{center}
	\caption{Diagram of the used main scale network $\mathcal{M}$ following a similar design of DSGN~\cite{chen2020dsgn}. A global voxel grid is used to sample image and cost volume features and construct a volumetric representation of the 3D scene in a pre-defined spatial range. This volumetric representation is converted into Bird's Eye View (BEV) feature maps. Object proposals are obtained by anchor classification and offset regression based on the BEV feature maps. Details are included in our SM.}
	\label{fig:mainscale}
\end{figure*}

\begin{figure*}
	\begin{center}
		\includegraphics[width=1\linewidth, trim=0cm 2cm 0cm 0cm]{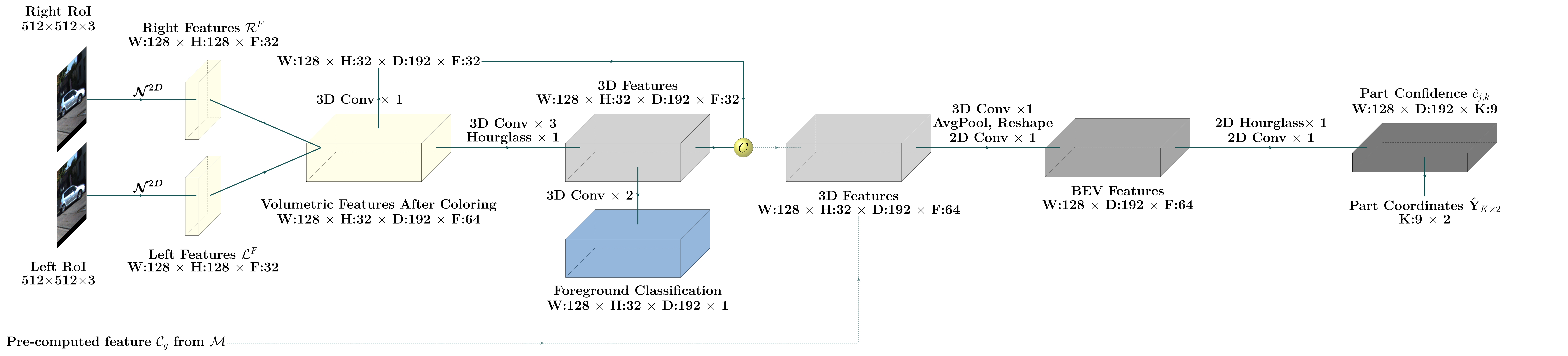}
	\end{center}
	\caption{Architecture of the Vernier scale network $\mathcal{V}$. High-resolution features are extracted from left/right regions of interest (RoIs). A uniform 3D grid conditioned on the current prediction is sampled. Each grid point is projected back to the RoIs to aggregate image features. Pre-computed features from $\mathcal{M}$ are used if available. The grid features are processed by a 3D CNN to predict the part confidence maps which implies the pose update.}
	\label{fig:vernierscale}
\end{figure*}

\noindent \textbf{Output representation.}
For each $p_q^c$, we encode the ground truth update as dense multi-part confidence since we implement $\mathcal{N}^{3D}$ as a CNN which is known good at dense classification tasks. For part $m$ ($m=1,\dots,K$), we assign each candidate a ground truth \emph{confidence} $c_{i,j,k}^m$. In implementation we adopt the ground plane assumption~\cite{oftnet} for autonomous driving datasets and ignore the offset in the height direction, which leads to a Bird's Eye View (BEV) confidence map as $c_{j,k}^m$. This does not reduce the generality of our framework and one can keep the original dimension if the height offset is significant in a different dataset. Denote the ground truth location for part $m$ as $(j^\star, k^\star)$. The confidence map is defined as 
\begin{equation}
c_{j,k}^m = e^{-\frac{(j - j^\star)^2 + (k - k^\star)^2}{\sigma^2}}. 
\end{equation} 
In total, $K$=9 parts including the ground truth cuboid center and its 8 corners are used as similarly defined in Eq.~(\ref{parts}). To reduce quantization errors, we transform the predicted confidences $\hat{c}_{j,k}$ into x-z coordinates $\hat{\text{Y}}_{K \times 2}$ using several convolution layers as in~\cite{Li_2021_CVPR}. 

\noindent \textbf{Model instantiation.}
We have specified the inputs and outputs of $\mathcal{V}$, where detailed parameters and network components $\mathcal{N}^{2D}$ and $\mathcal{N}^{3D}$ are abstracted away. This makes the framework flexible and one can specify these parameters and sub-networks based on its computation budget. In our study, $\mathcal{N}^{2D}$ is implemented as HRNet-W32~\cite{sun2019deep}. After coloring the voxel grid, we use 3D convolution layers and a 3D hourglass network to extract 3D spatial features. The 3D features are pooled and reshaped into BEV feature maps that are further transformed into confidence maps. The network architecture of $\mathcal{V}$ is depicted in Fig.~\ref{fig:vernierscale} and detailed in the SM. 

\section{Error-statistics-agnostic Training}
We train $\mathcal{M}$ and $\mathcal{V}$ separately. The training process of $\mathcal{M}$ can vary for different possible implementations. For the $\mathcal{M}$ used in our study, we use a similar procedure as the anchor-based baseline~\cite{chen2020dsgn}. The training supervision consists of a depth regression loss, an anchor classification loss, and an offset regression loss. The details are in the SM. 

Training $\mathcal{V}$ is the key step in our framework, where the inputs and targets for training $\mathcal{V}$ are not readily available. We propose a simple \emph{error-statistics-agnostic} (ESA) strategy to synthesize training data. This training strategy does not depend on $\mathcal{M}$ and can be used for any proposal model conveniently. Notably, PoseFix~\cite{moon2019posefix} proposed a refinement model for a different task and assumes there are certain types of errors from the predictions of the coarse models. Such error statistics are themselves collected from extra validation results. In contrast, our strategy assumes a Gaussian prior and does not require more specific knowledge of the error behavior.   

\noindent \textbf{Generating training data.}
For each ground truth 3D bounding box $\mathcal{B}_i = (x_i, y_i, z_i, h_i, w_i, l_i, \theta_i)$, we simulate a coarse prediction by adding a noise vector $n_i = (n^x_i, n^y_i, n^z_i, n^h_i, n^w_i, n^l_i, n^{\theta}_i)$ where $n \sim \text{N}(\textbf{0}, \Sigma)$. We further assume that the noise for each attribute is independent and the covariance matrix $\Sigma$ is diagonal. In experiments the standard derivations for the above attributes are 0.3m, 0m, 0.3m, 5cm, 5cm, 5cm and 5$^{\circ}$ respectively. Larger noise can be used if one assumes a weaker $\mathcal{M}$. The simulated coarse prediction $p_i^c = \mathcal{B}_i + n_i$, along with the ground truth confidence maps of $\mathcal{B}_i$, forms one training pair of $\mathcal{V}$. The Gaussian perturbation is added on-line for each instance in every iteration, which behaves like data augmentation so that $\mathcal{V}$ does not over-fit to a special subset of inputs.

\noindent \textbf{Loss function.}
We penalize the predicted confidence maps of $\mathcal{V}$ with $L_2$ loss as
$
L_{conf} = L_2(\hat{c}_{j,k}, c_{j,k}).
$ 
The transformed coordinates are penalized with smooth $L_1$ loss $L_{coord} = SL_{1}(\hat{\text{Y}}_{K \times 2}, \text{Y}_{K \times 2})$. For training $\mathcal{V}$-S models, we have $L_{total}^S = L_{conf} + L_{coord}$.

For $\mathcal{V}$-A models, apart from supervising these regression targets, we add an extra intermediate supervision since we cannot reuse features encoding the scene depth. We add a 3D convolution head that classifies for each candidate $d_{i,j,k}$ whether it is a foreground or not. This serves to add depth cues to train $\mathcal{V}$ similar to \cite{chen2020dsgn} yet different in that we directly add supervision in the 3D space instead of the cost volume used in \cite{chen2020dsgn}. For each point in the captured point cloud in $r_q$, the candidate it occupies after coordinate quantization is treated as a foreground. A candidate outside of the ground truth box $\mathcal{B}_q$ is treated as background. All other candidates are not assigned labels since they can be foreground (a foreground not recorded by the LiDAR) as well as background (free space). Since there are much more background candidates than the foreground ones, we use focal loss~\cite{lin2017focal} to supervise this classification task as 
\begin{align*}
L_{fg} & = \begin{cases}
-\alpha(1 - \hat{p}_{i,j,k})^\gamma \text{log}(\hat{p}_{i,j,k}), &p_{i,j,k}=1,\\
-(1 - \alpha)\hat{p}_{i,j,k}^\gamma \text{log}(1 - \hat{p}_{i,j,k}), & p_{i,j,k}=0\\
0, & \mbox{else}.
\end{cases}
\end{align*}
where $\hat{p}_{i,j,k}$ is the predicted foreground probability of candidate $d_{i,j,k}$ and $p_{i,j,k}$ is the ground-truth one where $p_{i,j,k}=1$ for foreground. 
We use the default parameters $\gamma=2$ and $\alpha=0.25$ and the total training loss $L_{total}^A = L_{conf} + L_{coord} + L_{fg}$. 

\begin{figure*}[t]
	\centering
	\includegraphics[width=1\linewidth, trim=0cm 0cm 0cm 0cm]{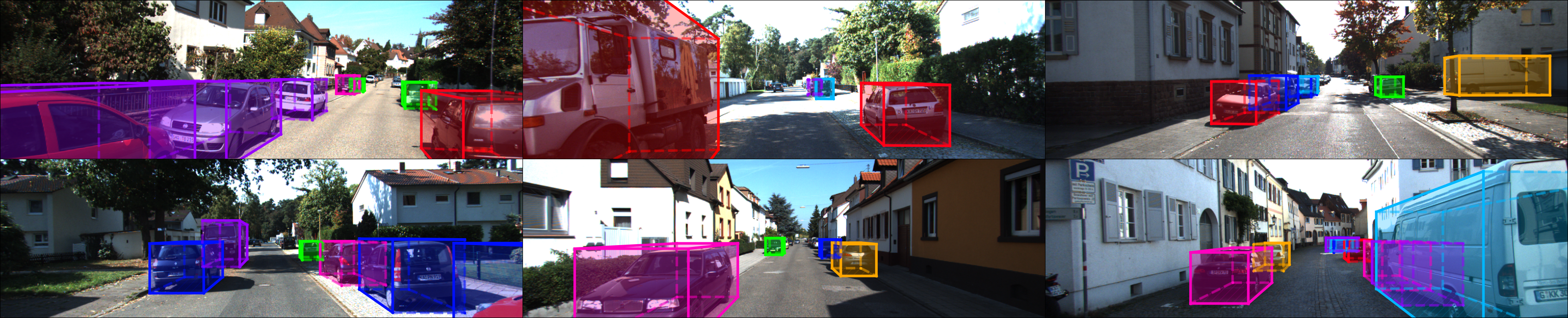}
	\caption{Qualitative results on KITTI \emph{val} split. Best viewed in color. More results can be found in the supplementary material.}	
	\label{fig:qualitative}
\end{figure*}

\begin{table*}
	\footnotesize
	\centering
	\begin{tabular}{|c|c|c|c|c|c|c|c|}
		\hline
		 &  & \multicolumn{3}{c|}{$AP_{3D}@R_{11}$} & \multicolumn{3}{c|}{$AP_{BEV}@R_{11}$} \\ \cline{3-8}
		\multirow{-2}{*}{Method}& \multirow{-2}{*}{Reference} &Easy & Moderate & Hard & Easy & Moderate & Hard\\ \hline 	
		MLF \cite{MLF} & CVPR' 18 & -- & 9.80 & -- & -- & 19.54 & -- \\  
		TLNet \cite{triangulation} &CVPR' 19 & 18.15 & 14.26 & 13.72 & 29.22 & 21.88 & 18.83 \\
		Stereo R-CNN \cite{stereorcnn} &CVPR' 19 & 54.1 & 36.7 & 31.1 & 68.5 & 48.3 & 41.5 \\ 
		PL: F-PointNet \cite{pseudolidar} &CVPR' 19 & 59.4 & 39.8 & 33.5 & 72.8 & 51.8 & 33.5  \\ 
		PL++: AVOD \cite{pseudo++} &ICLR' 19 & 63.2 & 46.8 & 39.8 & 77.0 & 63.7 & 56.0  \\		
		IDA-3D \cite{peng2020ida}  & CVPR' 20& 54.97 & 37.45 & 32.23 & 70.68 & 50.21 & 42.93 \\ 
		Disp-RCNN \cite{sun2020disp}  & CVPR' 20 & 64.29 & 47.73 & 40.11 & 77.63 & 64.38 & 50.68 \\ 
		DSGN \cite{chen2020dsgn} & CVPR' 20 & \textcolor{red}{72.31} & \textcolor{red}{54.27} & \textcolor{red}{47.71} & \textcolor{red}{83.24} & 63.91 & \textcolor{red}{57.83} \\ 
		ZoomNet \cite{xu2020zoomnet} & AAAI' 20 & 62.96 & 50.47 & 43.63 & 78.68 & \textcolor{red}{66.19} & 57.60 \\
		RTS-3D \cite{li2020rts3d} & AAAI' 21 & 64.76 & 46.70 & 39.27 & 77.50 & 58.65 & 50.14 \\ 
		Disp-RCNN-flb* \cite{chen2021shape}  & T-PAMI' 21 & 70.11
		&54.43 & 47.40 & 77.47 & 66.06 & 57.76 \\		
		\cline{1-8}
		SNVC (Ours) & AAAI' 22 & {\bf 77.29}  & {\bf 63.75}  & {\bf 56.81} &{\bf 87.07}  & {\bf72.95}  & {\bf66.77}\\
		\hline
	\end{tabular}
	\caption{Quantitative comparison of $AP_{3D}$ and $AP_{BEV}$ with SOTA stereo 3D object detection approaches on KITTI \emph{val} split. Second-best methods are marked by red color. Disp-RCNN-flb* requires extra mask annotation for training. 11 recall values are used to make the comparison consistent.}
	\label{table:validation}
\end{table*}

\section{Confidence-aware Robust Inference}
During inference, the update prediction $p_q^f$ derives from the predicted confidence maps $\hat{c}_{j,k}$ and x-z coordinates $\hat{\text{Y}}_{K \times 2}$, which indicate the tentative update position for each part. Certain study~\cite{peng2020ida} on instance-level S3DOD models only predicts the center of the object. This is similar to $K=1$ in our framework, where the predicted new center is used to refine the instance translation. However, this could lead to sub-optimal results when the center is occluded or hard to estimate. We thus propose an update strategy by utilizing $K$=9 parts along with predicted confidence. The predicted coordinates $\hat{\text{Y}}_{K \times 2}$ may not define parallel and orthogonal edges. We thus employ a 9-point registration approach by estimating a rigid transformation $\{\hat{\text{R}}, \hat{\text{T}}\}$ as
\begin{equation}
\label{transform}
\hat{\text{R}}, \hat{\text{T}} =\arg\min_{\text{R},\text{T}}\text{W}||\text{R}\tilde{{\text{Y}}}_{K \times 2} + \text{T} - \hat{\text{Y}}_{K \times 2}||,
\end{equation}
where $\tilde{{\text{Y}}}_{K \times 2}$ is the same part coordinates of the current proposal and W is the diagonal matrix with part confidences as its non-zero elements. 
The closed-form solution to Eq.~(\ref{transform}) is
\begin{equation}
\hat{\text{R}} = \text{V}^T \text{U}^T
\end{equation}
\begin{equation}
\hat{\text{T}} = -\hat{\text{R}} \tilde{{\text{Y}}}_{centroid} + \hat{{\text{Y}}}_{centroid}
\end{equation}
where U, V gives the singular decomposition as $\text{U}\text{S}\text{V} = \tilde{{\text{Y}}}^T\text{W}\hat{{\text{Y}}}$ where $\tilde{{\text{Y}}}_{centroid}$ is the average location of the K parts. This solution is the global optimum of Eq.~(\ref{transform}) as detailed in the SM. The refined 3D box is then obtained by applying this estimated transformation to the current proposal.

\section{Experiments}
We first introduce the used benchmark and evaluation metrics and then compare the overall performance of our MR SNVC with previously published approaches. We then demonstrate how our approach can improve existing 3D object detectors as a plug-and-play module. Finally, we present an ablation study on key design factors of $\mathcal{V}$.

\noindent\textbf{Dataset.} We employ the KITTI object detection benchmark~\cite{geiger2012we} for evaluation, which contains outdoor RGB images captured with calibrated stereo cameras. The dataset is split into 7,481 training images and 7,518 testing images. The training images are further split into the \emph{train} split and the \emph{val} split containing 3,712 and 3,769 images respectively. We use the \emph{train} split for training and conduct hyper-parameter tuning on the \emph{val} split. When reporting the model performance on the testing set, both the \emph{train} split and the \emph{val} split are used for training.

\noindent\textbf{Evaluation metrics.} We conduct the evaluation for the car category. We employ the official average precision metrics to validate our approach. \emph{3D Average Precision} ($AP_{3D}$) measures precision at 41 uniformly sampled recall values where a true positive is a predicted 3D box that has 3D intersection-over-union (IoU) $>$ 0.7 with the ground truth. \emph{Bird Eye's View Average Precision} ($AP_{BEV}$) instead uses 2D IoU $>$ 0.7 as the criterion where the 3D boxes are projected to the ground plane and the object heights are ignored. The KITTI benchmark further defines three sets of ground truth labels with different difficulty levels as easy, moderate, and hard. The difficulty level of one ground truth label is determined according to its 2D bounding box height, its occlusion level, and its truncation level. Evaluation is performed in parallel in these three different sets of ground truth labels. The hard set contains all the ground truth labels while the easy and moderate sets contain a fraction of easier objects. For consistency with previous works, results on the \emph{val} split are reported using 11 recall values (denoted as $@R_{11}$).

\noindent\textbf{Training details} are attached in our SM. 

\noindent\textbf{Comparison with state-of-the-arts.} Tab.~\ref{table:validation} and Tab.~\ref{table:test} compares the 3D object detection performance of our approach with other previously published methods on the \emph{val} split and the official testing set respectively. Fig.~\ref{fig:qualitative} shows 3D object detections of our system on the \emph{val} split. Our system outperforms previous approaches in all metrics with a clear margin. Compared with~\cite{chen2020dsgn} that only builds coarse scene-level representation, our system benefits from the proposed HR instance-level model that leads to more precise 3D localization. Compared with the approaches that build instance-level point cloud~\cite{xu2020zoomnet, sun2020disp}, our voxel-based representation shows superior performance, especially for the hard category. We believe the reason is that the instance point cloud in~\cite{xu2020zoomnet, sun2020disp} is sensitive to the depth estimation error and fails to handle distant and occluded objects due to a small number of points and lack of semantic visual features. In this camera-ready version paper, we also refer to a concurrent work~\cite{guo2021liga} starting from the same baseline (DSGN) as us which regularizes voxel representation learning by distilling knowledge from an extra pre-trained network. The contribution of distillation is complementary to ours. None of any methods in Tab.~\ref{table:validation}/\ref{table:test} needs this process known as learning from privileged information~\cite{lopez2015unifying} or an extra teacher.

\begin{table*}
	\footnotesize
	\centering
	\begin{tabular}{|c|c|c|c|c|c|c|c|}
		\hline
		\multirow{2}{*}{Method} & \multirow{2}{*}{Reference} & \multicolumn{3}{c|}{$AP_{3D}@R_{40}$} & \multicolumn{3}{c|}{$AP_{BEV}@R_{40}$} \\ \cline{3-8}
		& &Easy & Moderate & Hard & Easy & Moderate & Hard\\ \hline  
		TLNet \cite{triangulation} &CVPR' 19 & 7.64 & 4.37 & 3.74 & 13.71 & 7.69 & 6.73 \\
		Stereo R-CNN \cite{stereorcnn} &CVPR' 19 & 47.58 & 30.23 &23.72 & 61.92 & 41.31 & 33.42 \\ 
		ZoomNet \cite{xu2020zoomnet} & AAAI' 20 & 55.98 & 38.64 & 30.97 & 72.94 & 54.91 & 44.14 \\
		IDA-3D \cite{peng2020ida} & CVPR' 20 & 45.09 &29.32 &23.13 & 61.87 &42.47 &34.59 \\		
		Disp-RCNN \cite{sun2020disp} & CVPR' 20 & 59.58 &39.34 &31.99 & 74.07 &52.34 &43.77 \\
		DSGN \cite{chen2020dsgn} & CVPR' 20 & {73.50} & {52.18} & {45.14} & {82.90} & 65.05 & {56.60} \\ 
		CDN \cite{div2020wstereo} & NeurIPS' 20 & \textcolor{red}{74.52} &	\textcolor{red}{54.22} &	\textcolor{red}{46.36}  & \textcolor{red}{83.32} &	\textcolor{red}{66.24} &	\textcolor{red}{57.65} \\ 		
		RTS-3D \cite{li2020rts3d} & AAAI' 21 & 58.51 & 37.38 & 31.12 & 72.17 & 51.79 & 43.19 \\ 
		Disp-RCNN-flb* \cite{chen2021shape} & T-PAMI' 21 & 68.21 &45.78 &37.73 & 79.76 &58.62 &47.73 \\
		\cline{1-8}
		SNVC (Ours) & AAAI' 22 & {\bf78.54} &	{\bf61.34} &	{\bf54.23} &{\bf86.88} &{\bf	73.61} &	{\bf64.49}\\
		\hline
	\end{tabular}
	\caption{Overall system performance evaluated with $AP_{3D}$ and $AP_{BEV}$, and compared with SOTA stereo 3D object detection approaches on KITTI \emph{test} set (official KITTI leader-board). Second-best methods are marked by red color. The official recall values are used.}
	\label{table:test}
\end{table*}

\begin{figure}
	\centering
	\includegraphics[width=0.96\linewidth, trim=0cm 0.5cm 0cm 0cm]{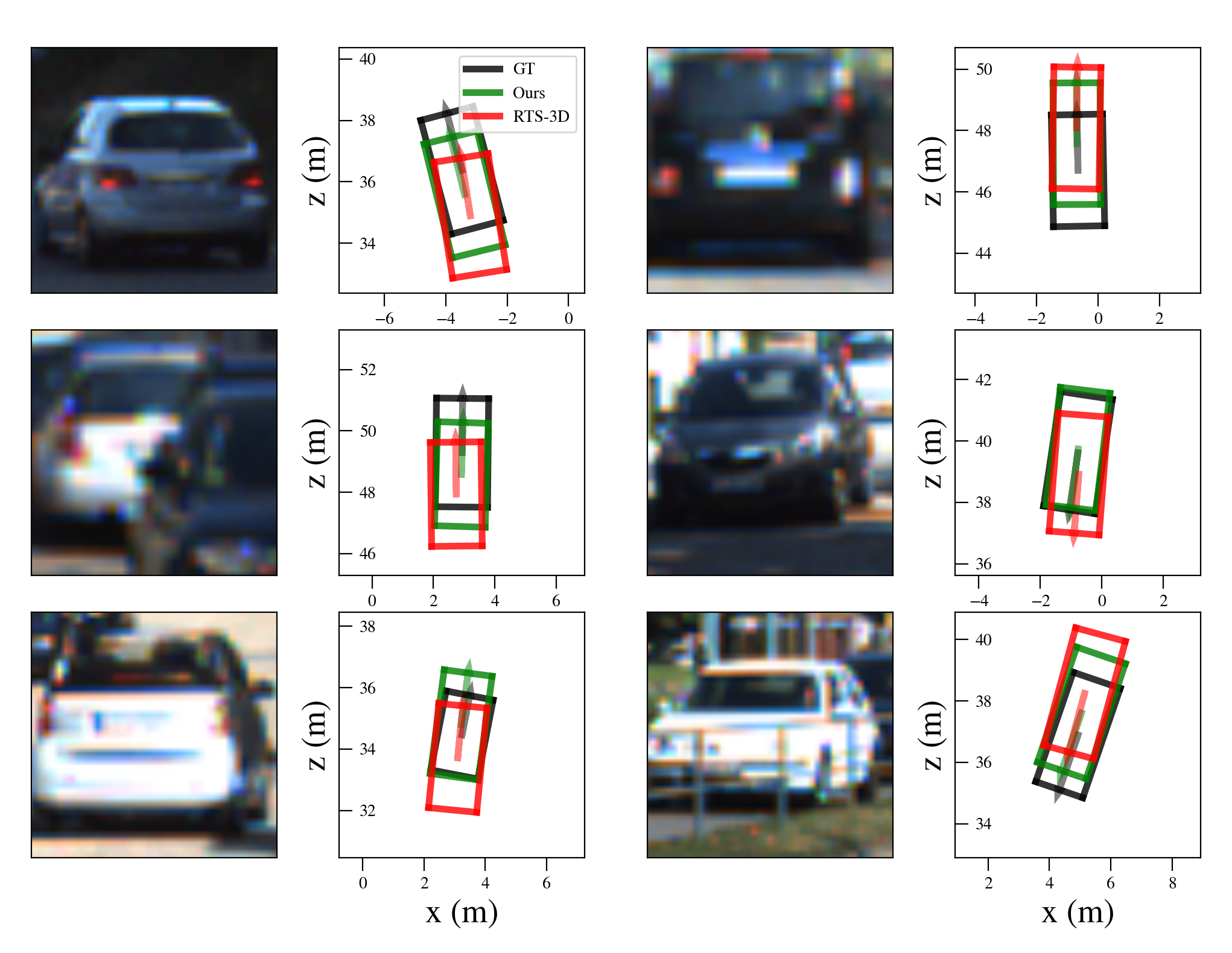}
	\caption{Qualitative comparison with RTS-3D~\cite{li2020rts3d} for objects on KITTI \emph{val} split. The image patches are RoIs on the left image and the 3D cuboid predictions are shown as bounding boxes in the bird's eye view plots. Note the objects are distant from the cameras.}	
	\label{quali-comp}
\end{figure}

\noindent\textbf{Model-agnostic refinement.} Tab.~\ref{tab:agnostic} shows the result when utilizing our $\mathcal{V}$ along with other existing 3D object detectors. We download the pre-trained weights from the official implementation\footnote{Different environments led to slight difference compared to the published results in Tab.~\ref{table:validation}.} of IDA-3D and RTS-3D to generate proposals and use our $\mathcal{V}$ to refine them. Note that using $\mathcal{V}$ leads to consistent and significant performance gain. This result validates that our $\mathcal{V}$ can be used as a model-agnostic refinement model. We show predictions for some distant and partially-occluded objects in Fig.~\ref{quali-comp} where our approach obtains better pose estimation performance compared to RTS-3D. RTS-3D can be used to generate coarse predictions in real-time applications while our $\mathcal{V}$ can complement it to obtain high-quality predictions only when necessary.

\begin{table}
	\footnotesize
	\centering
	\begin{tabular}{|l|c|c|c|}
		\hline
		\multirow{2}{*}{Method} & \multicolumn{3}{c|}{$AP_{3D}$/$AP_{BEV}@R_{11}$}\\ \cline{2-4}
		&  Easy & Moderate & Hard \\ 
		\hline
		RTS-3D &  59.31/74.49 & 41.61/53.98 & 34.67/46.66 \\
		RTS-3D + $\mathcal{V}$ &  69.25/82.71 & 52.92/65.75 & 45.82/57.47 \\
		IDA-3D  &  53.59/69.28 & 36.79/50.11 & 32.34/43.31 \\
		IDA-3D + $\mathcal{V}$ &64.13/80.81& 48.89/62.19& 43.00/55.03 \\
		\hline
	\end{tabular}
	\caption{$AP_{3D}/AP_{BEV}$ evaluated on the KITTI \emph{val} split when using $\mathcal{V}$-A as a model-agnostic refinement module with other 3D object detectors.}
	\label{tab:agnostic}
\end{table}

\begin{figure}
	\centering
	\includegraphics[width=1\linewidth, trim=0.5cm 0.5cm 0cm 0cm]{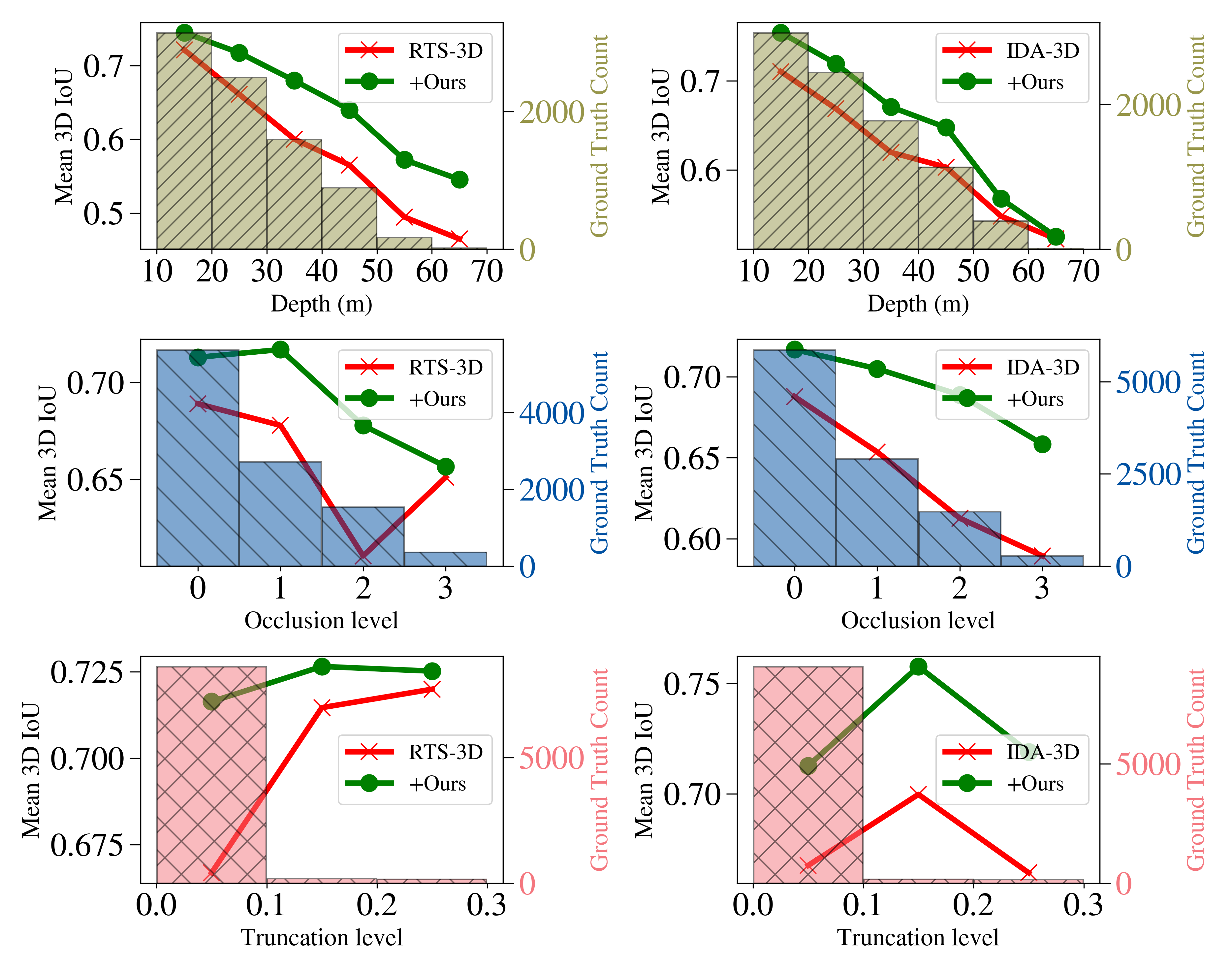}
	\caption{3D object detection performance with respect to ground truth depth (top), occlusion level (middle) and truncation level (bottom) on the KITTI \emph{val} split. Each bar indicates the number of ground truth (GT) objects that fall in the corresponding bin and are detected with 3D IoU $>$ 0.3. A red cross shows the average 3D IoU for those detected GT objects in a bin before utilizing our $\mathcal{V}$ for refinement. A green dot shows such a result after using $\mathcal{V}$. The occlusion level and truncation level are annotated for each ground truth object in the KITTI benchmark.}	
	\label{distribution}
\end{figure}

\noindent\textbf{Which objects benefit more from $\mathcal{V}?$} Tab.~\ref{tab:agnostic} shows the overall improvement in 3DOD performance, yet it does not reflect how the improvement is related to different object attributes. Fig.~\ref{distribution} instead shows which objects enjoy more performance improvements using the same proposals as in Tab.~\ref{tab:agnostic}. One can use such knowledge to decide which coarse proposal to refine in practice. Each ground truth (GT) object is assigned to the corresponding bin based on an attribute such as its depth, if there is one matching predicted object. A predicted object matches a ground truth if their 3D IoU $>$ 0.3. The match that has the largest 3D IoU is recorded for each GT object. The average of the matching 3D IoUs is shown as the line plots for GT objects in each bin. 

The detection quality in terms of $AP_{3D}$ improves for GT objects in all bins after using our $\mathcal{V}$ for refinement, in terms of all three attributes that influence the difficulty of detection. Note even if the objects are heavily occluded or partially truncated, using $\mathcal{V}$ still leads to a robust performance boost. RTS-3D also uses a voxel-based representation, and we can observe that the improvement over it for the middle-range and distant objects is more significant than the nearby objects. This validates our assumption that extracting complementary HR features are more helpful for the tiny objects that only have LR representations in the global feature maps.

\noindent\textbf{Effect of learning multiple object parts.} To validate our multi-part registration approach, we re-train another model to predict only the center confidence. This model uses the predicted center to update the proposal translation during inference. The performance comparison is shown in Tab.~\ref{tab:part-aware}, where our multi-part strategy leads to consistently better performance since it is more robust to partially-occluded objects whose visible parts can be used to provide a more reliable estimate.
\begin{table}
	\footnotesize
	\centering
	\begin{tabular}{|l|c|c|c|}
		\hline
		\multirow{2}{*}{Method} & \multicolumn{3}{c|}{$AP_{3D}$/$AP_{BEV}@R_{11}$}\\ \cline{2-4}
		&  Easy & Moderate & Hard \\ 
		\hline
		Center-only (K=1)   &64.94/77.89  & 47.22/64.00  & 43.73/55.84 \\
		Part-based (K=9)   &69.25/82.71 & 52.92/65.75 & 45.82/57.47 \\
		\hline
	\end{tabular}
	\caption{The same evaluation for RTS-3D + $\mathcal{V}$ as in Tab.~\ref{tab:agnostic} with varying number of parts K.}
	\label{tab:part-aware}
\end{table}

\noindent\textbf{Effect of voxel size.} To demonstrate the advantage of learning an HR voxel representation, we train $\mathcal{V}$ with a varying voxel resolution and the results of using these variants with RTS-3D are shown in Tab.~\ref{tab:voxel_size}. We can observe that using a smaller voxel size significantly improves the 3DOD performance which justifies our choice of parameters. While using a much coarser voxel grid (second row) leads to worse results, such performance is still better than using RTS-3D along without our $\mathcal{V}$. However, when the voxel grid is extremely coarse (first row), $\mathcal{V}$ cannot learn effective volumetric representation for S3DOD.

\begin{table}
	\footnotesize
	\centering
	\begin{tabular}{|l|c|c|c|}
		\hline
		\multirow{2}{*}{$(N_L, N_H, N_W)$} & \multicolumn{3}{c|}{$AP_{3D}$/$AP_{BEV}@R_{11}$}\\ \cline{2-4}
		&  Easy & Moderate & Hard \\ 
		\hline
		(24, 16, 16)  &  33.65/43.80 &27.84/37.38  &24.30/32.64 \\
		(48, 16, 32)    &60.10/73.25  & 44.31/56.19  & 37.45/53.30 \\
		(192, 32, 128)   &69.25/82.71 & 52.92/65.75 & 45.82/57.47 \\
		\hline
	\end{tabular}
	\caption{The same evaluation for RTS-3D + $\mathcal{V}$ as in Tab.~\ref{tab:agnostic} with varying voxel size.}
	\label{tab:voxel_size}
\end{table} 

\begin{figure}
	\centering
	\includegraphics[width=1\linewidth, trim=0cm 0.5cm 0cm 0cm]{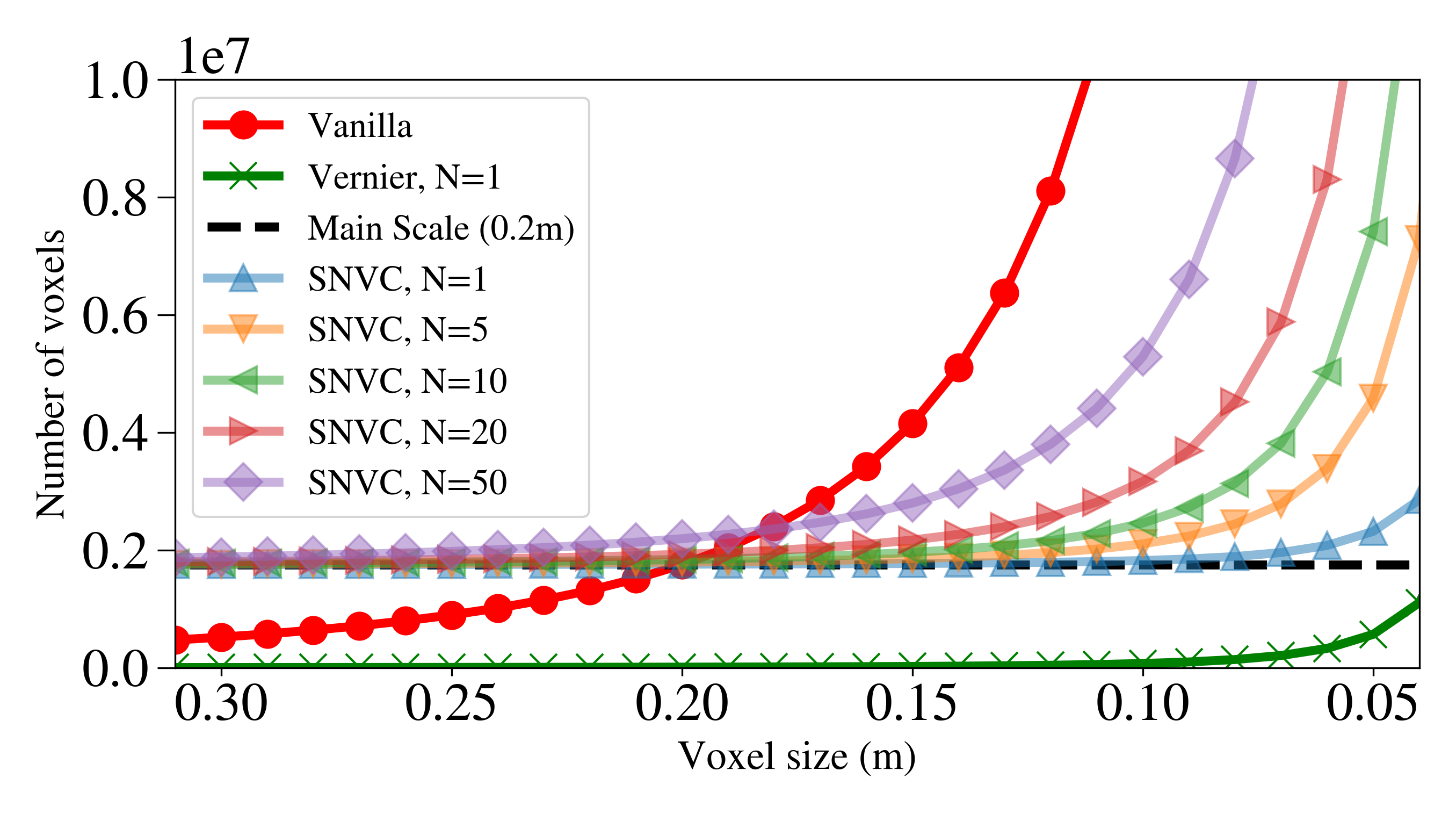}
	\caption{A comparison of needed number of voxels (NoVs) w and w/o our MR strategy. Vanilla: NoVs used for representing a global scene using a uniform resolution. Vernier: NoVs used to model a local scene in our $\mathcal{V}$. Main scale: NoVs used in our $\mathcal{M}$ with a resolution of 0.2m. SNVC, N=$x$: total NoVs as a sum of NoVs used in $\mathcal{M}$ plus NoVs used in building local scene for $x$ proposals.}	
	\label{memory}
\end{figure}

\noindent\textbf{Comparison of space requirement.} Building a uniform voxel grid for the global scene as used in the \emph{scene-centric} approaches needs intimidating memory that is not scalable to smaller voxel size. A comparison of the required number of voxels (NoVs) required in such systems and that needed in our proposed MR system is shown in Fig.~\ref{memory}. The NoVs used in the vanilla approach is $\mathcal{Q}(L_g, W_g, H_g, \Delta) = \frac{L_g * W_g * H_g}{\Delta^3}$ where $[L_g, W_g, H_g] = [\text{60m}, \text{60m}, \text{4m}]$ are the global spatial range and $\Delta$ is the voxel size. In contrast, the NoVs used in our approach is $N_L*N_H*N_W*N + \mathcal{Q}(L_g, W_g, H_g, \Delta_g)$ where $\Delta_g$ is the used global voxel size in the main scale network (0.2m). Note that our SNVC uses significantly fewer NoVs than the vanilla approach to achieve a finer representation beyond $\Delta_g$, which enables our framework to focus on important regions.

\noindent\textbf{More ablation studies and results} can be found in our SM.

\section{Conclusion} We introduce the idea of multi-resolution modeling to voxel-based stereo 3D object detection by modeling different regions with varying resolutions. This approach can keep the detection problem computationally tractable and can model important regions with smaller voxels to achieve high precision. A new instance-level model is designed, which samples candidate 3D locations and uses predicted object part coordinates to estimate a pose update. Our approach is validated to achieve state-of-the-art stereo 3D object detection performance and can perform model-agnostic refinement. For future study, instead of using a sampling grid with a fixed range, information of previous frames can be used to build a motion model that helps predict the future object location and provide a better clue on where to sample.
\clearpage
\bibliography{egbib}

\begin{thebibliography}{51}
\providecommand{\natexlab}[1]{#1}

\bibitem[{Andriluka, Roth, and Schiele(2008)}]{andriluka2008people}
Andriluka, M.; Roth, S.; and Schiele, B. 2008.
\newblock People-tracking-by-detection and people-detection-by-tracking.
\newblock In \emph{2008 IEEE Conference on computer vision and pattern
  recognition}, 1--8. IEEE.

\bibitem[{Bellotto et~al.(2009)Bellotto, Sommerlade, Benfold, Bibby, Reid,
  Roth, Fern{\'a}ndez, Van~Gool, and Gonzalez}]{bellotto2009distributed}
Bellotto, N.; Sommerlade, E.; Benfold, B.; Bibby, C.; Reid, I.; Roth, D.;
  Fern{\'a}ndez, C.; Van~Gool, L.; and Gonzalez, J. 2009.
\newblock A distributed camera system for multi-resolution surveillance.
\newblock In \emph{2009 Third ACM/IEEE International Conference on Distributed
  Smart Cameras (ICDSC)}, 1--8. IEEE.

\bibitem[{Blaha et~al.(2016)Blaha, Vogel, Richard, Wegner, Pock, and
  Schindler}]{blaha2016large}
Blaha, M.; Vogel, C.; Richard, A.; Wegner, J.~D.; Pock, T.; and Schindler, K.
  2016.
\newblock Large-scale semantic 3d reconstruction: an adaptive multi-resolution
  model for multi-class volumetric labeling.
\newblock In \emph{Proceedings of the IEEE Conference on Computer Vision and
  Pattern Recognition}, 3176--3184.

\bibitem[{Brazil and Liu(2019)}]{brazil2019m3d}
Brazil, G.; and Liu, X. 2019.
\newblock M3d-rpn: Monocular 3d region proposal network for object detection.
\newblock In \emph{Proceedings of the IEEE/CVF International Conference on
  Computer Vision}, 9287--9296.

\bibitem[{Chen et~al.(2021)Chen, Sun, Xie, Zhang, Shuai, Jiang, Zhang, Bao, and
  Zhou}]{chen2021shape}
Chen, L.; Sun, J.; Xie, Y.; Zhang, S.; Shuai, Q.; Jiang, Q.; Zhang, G.; Bao,
  H.; and Zhou, X. 2021.
\newblock Shape Prior Guided Instance Disparity Estimation for 3D Object
  Detection.
\newblock \emph{IEEE Transactions on Pattern Analysis and Machine
  Intelligence}.

\bibitem[{Chen et~al.(2015)Chen, Kundu, Zhu, Berneshawi, Ma, Fidler, and
  Urtasun}]{3dop}
Chen, X.; Kundu, K.; Zhu, Y.; Berneshawi, A.~G.; Ma, H.; Fidler, S.; and
  Urtasun, R. 2015.
\newblock 3d object proposals for accurate object class detection.
\newblock In \emph{Advances in Neural Information Processing Systems},
  424--432. Citeseer.

\bibitem[{Chen et~al.(2017)Chen, Ma, Wan, Li, and Xia}]{MV3D}
Chen, X.; Ma, H.; Wan, J.; Li, B.; and Xia, T. 2017.
\newblock Multi-view 3d object detection network for autonomous driving.
\newblock In \emph{Proceedings of the IEEE conference on Computer Vision and
  Pattern Recognition}, 1907--1915.

\bibitem[{Chen et~al.(2020)Chen, Liu, Shen, and Jia}]{chen2020dsgn}
Chen, Y.; Liu, S.; Shen, X.; and Jia, J. 2020.
\newblock Dsgn: Deep stereo geometry network for 3d object detection.
\newblock In \emph{Proceedings of the IEEE/CVF Conference on Computer Vision
  and Pattern Recognition}, 12536--12545.

\bibitem[{Cheng et~al.(2020)Cheng, Xiao, Wang, Shi, Huang, and
  Zhang}]{cheng2020higherhrnet}
Cheng, B.; Xiao, B.; Wang, J.; Shi, H.; Huang, T.~S.; and Zhang, L. 2020.
\newblock Higherhrnet: Scale-aware representation learning for bottom-up human
  pose estimation.
\newblock In \emph{Proceedings of the IEEE/CVF Conference on Computer Vision
  and Pattern Recognition}, 5386--5395.

\bibitem[{Choy et~al.(2016)Choy, Xu, Gwak, Chen, and Savarese}]{choy20163d}
Choy, C.~B.; Xu, D.; Gwak, J.; Chen, K.; and Savarese, S. 2016.
\newblock 3d-r2n2: A unified approach for single and multi-view 3d object
  reconstruction.
\newblock In \emph{European conference on computer vision}, 628--644. Springer.

\bibitem[{Garg et~al.(2020)Garg, Wang, Hariharan, Campbell, Weinberger, and
  Chao}]{div2020wstereo}
Garg, D.; Wang, Y.; Hariharan, B.; Campbell, M.; Weinberger, K.~Q.; and Chao,
  W.-L. 2020.
\newblock Wasserstein Distances for Stereo Disparity Estimation.
\newblock In Larochelle, H.; Ranzato, M.; Hadsell, R.; Balcan, M.~F.; and Lin,
  H., eds., \emph{Advances in Neural Information Processing Systems},
  volume~33, 22517--22529. Curran Associates, Inc.

\bibitem[{Geiger, Lenz, and Urtasun(2012)}]{geiger2012we}
Geiger, A.; Lenz, P.; and Urtasun, R. 2012.
\newblock Are we ready for autonomous driving? the kitti vision benchmark
  suite.
\newblock In \emph{Proceedings of the IEEE/CVF Conference on Computer Vision
  and Pattern Recognition}, 3354--3361. IEEE.

\bibitem[{Guo et~al.(2021)Guo, Shi, Wang, and Li}]{guo2021liga}
Guo, X.; Shi, S.; Wang, X.; and Li, H. 2021.
\newblock Liga-stereo: Learning lidar geometry aware representations for
  stereo-based 3d detector.
\newblock In \emph{Proceedings of the IEEE/CVF International Conference on
  Computer Vision}, 3153--3163.

\bibitem[{Ke et~al.(2020)Ke, Li, Sun, Tai, and Tang}]{ke2020gsnet}
Ke, L.; Li, S.; Sun, Y.; Tai, Y.-W.; and Tang, C.-K. 2020.
\newblock Gsnet: Joint vehicle pose and shape reconstruction with geometrical
  and scene-aware supervision.
\newblock In \emph{European Conference on Computer Vision}, 515--532. Springer.

\bibitem[{Kundu, Li, and Rehg(2018)}]{kundu20183d}
Kundu, A.; Li, Y.; and Rehg, J.~M. 2018.
\newblock 3d-rcnn: Instance-level 3d object reconstruction via
  render-and-compare.
\newblock In \emph{Proceedings of the IEEE conference on computer vision and
  pattern recognition}, 3559--3568.

\bibitem[{Laine and Karras(2010)}]{laine2010efficient}
Laine, S.; and Karras, T. 2010.
\newblock Efficient sparse voxel octrees.
\newblock \emph{IEEE Transactions on Visualization and Computer Graphics},
  17(8): 1048--1059.

\bibitem[{Lang et~al.(2019)Lang, Vora, Caesar, Zhou, Yang, and
  Beijbom}]{lang2019pointpillars}
Lang, A.~H.; Vora, S.; Caesar, H.; Zhou, L.; Yang, J.; and Beijbom, O. 2019.
\newblock Pointpillars: Fast encoders for object detection from point clouds.
\newblock In \emph{Proceedings of the IEEE/CVF Conference on Computer Vision
  and Pattern Recognition}, 12697--12705.

\bibitem[{Li, Chen, and Shen(2019)}]{stereorcnn}
Li, P.; Chen, X.; and Shen, S. 2019.
\newblock Stereo r-cnn based 3d object detection for autonomous driving.
\newblock In \emph{Proceedings of the IEEE/CVF Conference on Computer Vision
  and Pattern Recognition}, 7644--7652.

\bibitem[{Li, Su, and Zhao(2021)}]{li2020rts3d}
Li, P.; Su, S.; and Zhao, H. 2021.
\newblock RTS3D: Real-time Stereo 3D Detection from 4D Feature-Consistency
  Embedding Space for Autonomous Driving.
\newblock In \emph{Proceedings of the AAAI Conference on Artificial
  Intelligence}, volume~35, 1930--1939.

\bibitem[{Li et~al.(2020)Li, Ke, Pratama, Tai, Tang, and
  Cheng}]{li2020cascaded}
Li, S.; Ke, L.; Pratama, K.; Tai, Y.-W.; Tang, C.-K.; and Cheng, K.-T. 2020.
\newblock Cascaded deep monocular 3D human pose estimation with evolutionary
  training data.
\newblock In \emph{Proceedings of the IEEE/CVF Conference on Computer Vision
  and Pattern Recognition}, 6173--6183.

\bibitem[{Li et~al.(2021)Li, Yan, Li, and Cheng}]{Li_2021_CVPR}
Li, S.; Yan, Z.; Li, H.; and Cheng, K.-T. 2021.
\newblock Exploring intermediate representation for monocular vehicle pose
  estimation.
\newblock In \emph{Proceedings of the IEEE/CVF Conference on Computer Vision
  and Pattern Recognition}, 1873--1883.

\bibitem[{Li, Wang, and Wang(2021)}]{li2021lidar}
Li, Z.; Wang, F.; and Wang, N. 2021.
\newblock LiDAR R-CNN: An Efficient and Universal 3D Object Detector.
\newblock In \emph{Proceedings of the IEEE/CVF Conference on Computer Vision
  and Pattern Recognition}, 7546--7555.

\bibitem[{Liang et~al.(2019)Liang, Yang, Chen, Hu, and Urtasun}]{MMF}
Liang, M.; Yang, B.; Chen, Y.; Hu, R.; and Urtasun, R. 2019.
\newblock Multi-task multi-sensor fusion for 3d object detection.
\newblock In \emph{Proceedings of the IEEE/CVF Conference on Computer Vision
  and Pattern Recognition}, 7345--7353.

\bibitem[{Lin et~al.(2017)Lin, Goyal, Girshick, He, and
  Doll{\'a}r}]{lin2017focal}
Lin, T.-Y.; Goyal, P.; Girshick, R.; He, K.; and Doll{\'a}r, P. 2017.
\newblock Focal loss for dense object detection.
\newblock In \emph{Proceedings of the IEEE international conference on computer
  vision}, 2980--2988.

\bibitem[{Liu et~al.(2019)Liu, Lu, Xu, Tian, and Zhou}]{liu2019deep}
Liu, L.; Lu, J.; Xu, C.; Tian, Q.; and Zhou, J. 2019.
\newblock Deep fitting degree scoring network for monocular 3d object
  detection.
\newblock In \emph{Proceedings of the IEEE/CVF Conference on Computer Vision
  and Pattern Recognition}, 1057--1066.

\bibitem[{Liu et~al.(2020)Liu, Wu, Lu, Xie, Zhou, and Tian}]{liu2020reinforced}
Liu, L.; Wu, C.; Lu, J.; Xie, L.; Zhou, J.; and Tian, Q. 2020.
\newblock Reinforced axial refinement network for monocular 3d object
  detection.
\newblock In \emph{European Conference on Computer Vision}, 540--556. Springer.

\bibitem[{Lopez-Paz et~al.(2015)Lopez-Paz, Bottou, Sch{\"o}lkopf, and
  Vapnik}]{lopez2015unifying}
Lopez-Paz, D.; Bottou, L.; Sch{\"o}lkopf, B.; and Vapnik, V. 2015.
\newblock Unifying distillation and privileged information.
\newblock \emph{arXiv preprint arXiv:1511.03643}.

\bibitem[{Lu et~al.(2021)Lu, Ma, Yang, Zhang, Liu, Chu, Yan, and
  Ouyang}]{lu2021geometry}
Lu, Y.; Ma, X.; Yang, L.; Zhang, T.; Liu, Y.; Chu, Q.; Yan, J.; and Ouyang, W.
  2021.
\newblock Geometry uncertainty projection network for monocular 3d object
  detection.
\newblock In \emph{Proceedings of the IEEE/CVF International Conference on
  Computer Vision}, 3111--3121.

\bibitem[{Moon, Chang, and Lee(2019)}]{moon2019posefix}
Moon, G.; Chang, J.~Y.; and Lee, K.~M. 2019.
\newblock Posefix: Model-agnostic general human pose refinement network.
\newblock In \emph{Proceedings of the IEEE/CVF Conference on Computer Vision
  and Pattern Recognition}, 7773--7781.

\bibitem[{Peng et~al.(2020)Peng, Pan, Liu, and Sun}]{peng2020ida}
Peng, W.; Pan, H.; Liu, H.; and Sun, Y. 2020.
\newblock Ida-3d: Instance-depth-aware 3d object detection from stereo vision
  for autonomous driving.
\newblock In \emph{Proceedings of the IEEE/CVF Conference on Computer Vision
  and Pattern Recognition}, 13015--13024.

\bibitem[{Qi et~al.(2018)Qi, Liu, Wu, Su, and Guibas}]{fpointnet}
Qi, C.~R.; Liu, W.; Wu, C.; Su, H.; and Guibas, L.~J. 2018.
\newblock Frustum pointnets for 3d object detection from rgb-d data.
\newblock In \emph{Proceedings of the IEEE conference on computer vision and
  pattern recognition}, 918--927.

\bibitem[{Qin, Wang, and Lu(2019)}]{triangulation}
Qin, Z.; Wang, J.; and Lu, Y. 2019.
\newblock Triangulation learning network: from monocular to stereo 3d object
  detection.
\newblock In \emph{Proceedings of the IEEE/CVF Conference on Computer Vision
  and Pattern Recognition}, 7615--7623.

\bibitem[{Reading et~al.(2021)Reading, Harakeh, Chae, and
  Waslander}]{reading2021categorical}
Reading, C.; Harakeh, A.; Chae, J.; and Waslander, S.~L. 2021.
\newblock Categorical depth distribution network for monocular 3d object
  detection.
\newblock In \emph{Proceedings of the IEEE/CVF Conference on Computer Vision
  and Pattern Recognition}, 8555--8564.

\bibitem[{Riegler, Osman~Ulusoy, and Geiger(2017)}]{riegler2017octnet}
Riegler, G.; Osman~Ulusoy, A.; and Geiger, A. 2017.
\newblock Octnet: Learning deep 3d representations at high resolutions.
\newblock In \emph{Proceedings of the IEEE conference on computer vision and
  pattern recognition}, 3577--3586.

\bibitem[{Roddick, Kendall, and Cipolla(2019)}]{oftnet}
Roddick, T.; Kendall, A.; and Cipolla, R. 2019.
\newblock Orthographic Feature Transform for Monocular 3D Object Detection.
\newblock In Sidorov, K.; and Hicks, Y., eds., \emph{Proceedings of the British
  Machine Vision Conference (BMVC)}, 59.1--59.13. BMVA Press.

\bibitem[{Seitz and Dyer(1999)}]{seitz1999photorealistic}
Seitz, S.~M.; and Dyer, C.~R. 1999.
\newblock Photorealistic scene reconstruction by voxel coloring.
\newblock \emph{International Journal of Computer Vision}, 35(2): 151--173.

\bibitem[{Shi, Wang, and Li(2019)}]{shi2019pointrcnn}
Shi, S.; Wang, X.; and Li, H. 2019.
\newblock Pointrcnn: 3d object proposal generation and detection from point
  cloud.
\newblock In \emph{Proceedings of the IEEE/CVF conference on computer vision
  and pattern recognition}, 770--779.

\bibitem[{Snow, Viola, and Zabih(2000)}]{snow2000exact}
Snow, D.; Viola, P.; and Zabih, R. 2000.
\newblock Exact voxel occupancy with graph cuts.
\newblock In \emph{Proceedings IEEE Conference on Computer Vision and Pattern
  Recognition. CVPR 2000 (Cat. No. PR00662)}, volume~1, 345--352. IEEE.

\bibitem[{Sun et~al.(2020)Sun, Chen, Xie, Zhang, Jiang, Zhou, and
  Bao}]{sun2020disp}
Sun, J.; Chen, L.; Xie, Y.; Zhang, S.; Jiang, Q.; Zhou, X.; and Bao, H. 2020.
\newblock Disp r-cnn: Stereo 3d object detection via shape prior guided
  instance disparity estimation.
\newblock In \emph{Proceedings of the IEEE/CVF Conference on Computer Vision
  and Pattern Recognition}, 10548--10557.

\bibitem[{Sun et~al.(2019)Sun, Xiao, Liu, and Wang}]{sun2019deep}
Sun, K.; Xiao, B.; Liu, D.; and Wang, J. 2019.
\newblock Deep high-resolution representation learning for human pose
  estimation.
\newblock In \emph{Proceedings of the IEEE/CVF Conference on Computer Vision
  and Pattern Recognition}, 5693--5703.

\bibitem[{Vogiatzis, Torr, and Cipolla(2005)}]{vogiatzis2005multi}
Vogiatzis, G.; Torr, P.~H.; and Cipolla, R. 2005.
\newblock Multi-view stereo via volumetric graph-cuts.
\newblock In \emph{2005 IEEE Computer Society Conference on Computer Vision and
  Pattern Recognition (CVPR'05)}, volume~2, 391--398. IEEE.

\bibitem[{Wang et~al.(2020)Wang, Sun, Cheng, Jiang, Deng, Zhao, Liu, Mu, Tan,
  Wang et~al.}]{wang2020deep}
Wang, J.; Sun, K.; Cheng, T.; Jiang, B.; Deng, C.; Zhao, Y.; Liu, D.; Mu, Y.;
  Tan, M.; Wang, X.; et~al. 2020.
\newblock Deep high-resolution representation learning for visual recognition.
\newblock \emph{IEEE transactions on pattern analysis and machine
  intelligence}.

\bibitem[{Wang et~al.(2021{\natexlab{a}})Wang, Du, Ye, Fu, Guo, Xue, Feng, and
  Zhang}]{wang2021depth}
Wang, L.; Du, L.; Ye, X.; Fu, Y.; Guo, G.; Xue, X.; Feng, J.; and Zhang, L.
  2021{\natexlab{a}}.
\newblock Depth-conditioned Dynamic Message Propagation for Monocular 3D Object
  Detection.
\newblock In \emph{Proceedings of the IEEE/CVF Conference on Computer Vision
  and Pattern Recognition}, 454--463.

\bibitem[{Wang et~al.(2019)Wang, Chao, Garg, Hariharan, Campbell, and
  Weinberger}]{pseudolidar}
Wang, Y.; Chao, W.-L.; Garg, D.; Hariharan, B.; Campbell, M.; and Weinberger,
  K.~Q. 2019.
\newblock Pseudo-lidar from visual depth estimation: Bridging the gap in 3d
  object detection for autonomous driving.
\newblock In \emph{Proceedings of the IEEE/CVF Conference on Computer Vision
  and Pattern Recognition}, 8445--8453.

\bibitem[{Wang et~al.(2021{\natexlab{b}})Wang, Yang, Hu, Liang, and
  Urtasun}]{wang2021plume}
Wang, Y.; Yang, B.; Hu, R.; Liang, M.; and Urtasun, R. 2021{\natexlab{b}}.
\newblock PLUME: Efficient 3D Object Detection from Stereo Images.
\newblock \emph{arXiv preprint arXiv:2101.06594}.

\bibitem[{Weng and Kitani(2019)}]{monopseudoliar}
Weng, X.; and Kitani, K. 2019.
\newblock Monocular 3d object detection with pseudo-lidar point cloud.
\newblock In \emph{Proceedings of the IEEE/CVF International Conference on
  Computer Vision Workshops}, 0--0.

\bibitem[{Xu and Chen(2018)}]{MLF}
Xu, B.; and Chen, Z. 2018.
\newblock Multi-level fusion based 3d object detection from monocular images.
\newblock In \emph{Proceedings of the IEEE conference on computer vision and
  pattern recognition}, 2345--2353.

\bibitem[{Xu et~al.(2020)Xu, Zhang, Ye, Tan, Yang, Wen, Ding, Meng, and
  Huang}]{xu2020zoomnet}
Xu, Z.; Zhang, W.; Ye, X.; Tan, X.; Yang, W.; Wen, S.; Ding, E.; Meng, A.; and
  Huang, L. 2020.
\newblock Zoomnet: Part-aware adaptive zooming neural network for 3d object
  detection.
\newblock In \emph{Proceedings of the AAAI Conference on Artificial
  Intelligence}, volume~34, 12557--12564.

\bibitem[{Yan, Mao, and Li(2018)}]{second}
Yan, Y.; Mao, Y.; and Li, B. 2018.
\newblock SECOND: Sparsely Embedded Convolutional Detection.

\bibitem[{You et~al.(2019)You, Wang, Chao, Garg, Pleiss, Hariharan, Campbell,
  and Weinberger}]{pseudo++}
You, Y.; Wang, Y.; Chao, W.-L.; Garg, D.; Pleiss, G.; Hariharan, B.; Campbell,
  M.; and Weinberger, K.~Q. 2019.
\newblock Pseudo-LiDAR++: Accurate Depth for 3D Object Detection in Autonomous
  Driving.
\newblock In \emph{ICLR}.

\bibitem[{Zhou and Tuzel(2018)}]{VoxelNet}
Zhou, Y.; and Tuzel, O. 2018.
\newblock Voxelnet: End-to-end learning for point cloud based 3d object
  detection.
\newblock In \emph{Proceedings of the IEEE conference on computer vision and
  pattern recognition}, 4490--4499.

\end{thebibliography}
\noindent\textbf{Acknowledgments} This work is supported by
Hong Kong Research Grants Council (RGC) General Research Fund (GRF) 16203319.
\end{document}